\documentclass[11pt,a4paper]{article}
\usepackage[hyperref]{naaclhlt2019}
\usepackage{times}
\usepackage{latexsym}
\usepackage{framed}

\aclfinalcopy 
\def\aclpaperid{11} 

\usepackage{graphicx}  
\usepackage{colortbl}
\usepackage{breqn}
\usepackage{amssymb}
\usepackage{multirow}
\usepackage{hyperref}

\newcommand{\scite}[1]{\citeauthor{#1} \shortcite{#1}}

\renewcommand{\url}[1]{\href{#1}{#1}}

\title{Towards Automatic Generation of Shareable Synthetic Clinical Notes \\ Using Neural Language Models}

\author{Oren Melamud \\
  IBM T. J. Watson Research Center  \\
  Yorktown Heights, NY, USA. \\
  {\tt oren.melamud@ibm.com} \\\And
  Chaitanya Shivade \\
  IBM Almaden Research Center \\
  San Jose, CA, USA. \\
  {\tt cshivade@us.ibm.com} \\}

\date{}

\begin{document}

\maketitle

\begin{abstract}
Large-scale clinical data is invaluable to driving many computational scientific advances today. However, understandable concerns regarding patient privacy hinder the open dissemination of such data and give rise to suboptimal siloed research. 
De-identification methods attempt to address these concerns but were shown to be susceptible to adversarial attacks.
In this work, we focus on the vast amounts of unstructured natural language data stored in clinical notes and propose to automatically generate synthetic clinical notes that are more amenable to sharing using generative models trained on real de-identified records. 
To evaluate the merit of such notes, we measure both their privacy preservation properties as well as utility in training clinical NLP models. Experiments using neural language models yield notes whose utility is close to that of the real ones in some clinical NLP tasks,
yet leave ample room for future improvements.
\end{abstract}

\section{Introduction}

Clinical data and clinical notes specifically, are an important factor for the advancement of computational methods in the medical domain. Suffice to say that the recently introduced \emph{MIMIC-III} clinical database alone \cite{johnson2016mimic} already has hundreds of cites on Google Scholar. However, understandable privacy concerns yield strict restrictions on clinical data dissemination, thus inhibiting scientific progress. 
De-identification techniques provide some relief \cite{dernoncourt2017identification}, but are still far from providing the privacy guarantees required for unrestricted sharing \cite{ohm2009broken,shokri2017membership}.

In this work, we 
investigate the possibility of disseminating
clinical notes data by computationally generating synthetic notes that are safer to share than real ones.
To this end, we introduce a clinical notes generation task, where synthetic notes are to be generated based on a set of real de-identified clinical discharge summary notes, henceforth referred to as \emph{MedText}, which we extracted from MIMIC-III. The evaluation includes a new measure of the privacy preservation properties of the synthetic notes, as well as their utility on three clinical NLP tasks. We use neural language models to perform this task and discuss the potential and challenges of this approach. Resources associated with this paper are available for download.
\footnote{\url{https://github.com/orenmel/synth-clinical-notes}}

\section{Background}

\subsection{Clinical Notes}
Electronic health records contain a wealth of information about patients in the form of both structured data 
and unstructured text. While structured data 
is critical for purposes like billing and administration, unstructured clinical notes contain important information entered by doctors, nurses, and other staff associated with patient care, which is not captured elsewhere. To this end, researchers have found that although structured data is easily accessible, clinical notes remain indispensable for understanding a patient record \cite{birman2005accuracy,singh2004accuracy}.
\scite{rosenbloom2011data}
argued that clinical notes are considered to be more useful for identifying patients with specific disorders.
A study by \scite{kopcke2013evaluation} found that 65\% of the data required to determine eligibility of a patient into clinical trials was not found in structured data and required examination of clinical notes. Similar findings were also reported by \scite{raghavan2014essential}. 

Due to their importance, it is no wonder that clinical notes are used extensively in medical NLP research. Unfortunately, however, due to privacy concerns, explained further below, it is very common that the data is exclusively available only to researchers collaborating with or working for a particular healthcare provider \cite{choi2016doctor,afzal2018natural,liu2018deep}.

\subsection{De-identification}
Clinical notes contain sensitive personal information required for medical investigations, which is protected by law. For example, in the United States, the Health Insurance Portability and Accountability Act (HIPAA)\footnote{Office for Civil Rights H. Standards for privacy of individually identifiable health information. Final rule. Federal Register.  2002;67:53181.} defines 18 types of protected health information (PHI) that needs to be removed to de-identify clinical notes (e.g. name, age, dates and contact details). Both manual and automated methods for de-identification have been investigated with varying degrees of success. 
\scite{neamatullah2008automated} reported a recall ranging from 0.63 to 0.94 between 14 clinicians for manually identifying PHI in 130 clinical notes.  Since human annotations for clinical data are costly \cite{douglass2004computer}, researchers have investigated automated and semi-automated methods for de-identification \cite{gobbel2014assisted,hanauer2013bootstrapping}. Automated methods range from rule-based systems \cite{morrison2009repurposing} to statistical methods such as support vector machines and conditional random fields \cite{stubbs2015automated}, with more recent use of recurrent neural networks \cite{liu2017identification,dernoncourt2017identification}. 

Unfortunately, despite strong results reported for
clinical data de-identification methods,
it is usually hard to determine to what extent they are resistant to re-identification attacks on healthcare data \cite{ohm2009broken,el2011systematic,gkoulalas2014publishing}. Therefore, in practice, de-identified patient data is almost never shared freely, and complementary privacy protection techniques, such as the one described in the following section, are being actively investigated.

\subsection{Differential Privacy}

Collections of private individual data records are commonly used to compute aggregated statistical information or train statistical models that are made  publicly available. Possible use cases include collections of search queries used to provide intelligent auto-completion suggestions to users of search engines
and medical records used to train computer-based clinical expert systems. 
While this is not always transparent, providing access to such aggregated information may be sufficient for attackers to infer some individual private data. One example to such well crafted attacks are the \emph{membership inference} attacks proposed by \scite{shokri2017membership}. In these attacks, the adversary has only black-box access to a machine learning model that was trained on a collection of records, and tries to learn how to infer whether any given data record was part of that model's train set or not. Susceptibility to such attacks is an indication that private information may be compromised.

Differential privacy (DP) is, broadly speaking, a guarantee that the personal information of each individual record within a collection is reasonably protected even when the aggregated statistical information is exposed. A model that is trained on some record collection as its input and makes its outputs publicly available, will provide stronger DP guarantees the less those outputs depend on the presence of any individual record in the collection.

More formally, 
a randomized function $K$ 
provides $\epsilon-$differential privacy if for all
collections $C_1$ and $C_2$ differing by at most one element, and all $S \subseteq Range(K)$:

$$
\log p(K(C_1) \in S) - \log p(K(C_2) \in S) \leq \epsilon 
$$

A mechanism $K$ satisfying this definition addresses concerns of personal information leakage from any individual record since 
the inclusion of that record will not result in any publicly exposed outputs becoming significantly more or less likely \cite{dwork2008differential}.

Differential privacy is an active research field, with various techniques proposed to provide DP guarantees to various machine learning models \cite{abadi2016deep,papernot2018scalable}. However, while DP shares some motivation with traditional machine learning techniques, such as the need to avoid overfitting, it is unfortunately not always easy to achieve good differential privacy guarantees, and they typically come at the cost of some accuracy degradation and computational complexity.

\subsection{Language Modeling}

Language models (LMs) learn to estimate the probability of a next word given a context of preceding words, i.e. $\hat{P}(w_i|w_{1..i-1})$, where $w_i$ is the word in position $i$ in the text. 
They were found useful in many NLP tasks, including text classification \cite{howard2018fine}, machine translation \cite{luong2015deep} and speech recognition \cite{chen2015recurrent}.
They are also commonly used for generating text \cite{sutskever2011generating,radford2018language} as we do in this paper. To generate text, a trained model is typically used to estimate the conditional probability distribution of the next word $\hat{P}(w_i|w_{1..i-1})$. Next, it samples a word for position~$i$ from this distribution and then goes on to sample the next one based on $\hat{P}(w_{i+1}|w_{1..i})$ and so on. The predominant model design used to implement LMs today used to be Recurrent Neural Networks (RNNs) due to their ability to capture long distance contexts \cite{jozefowicz2016exploring}, but recently, the attention-based \emph{Transformer} architecture surpassed state of the art results \cite{radford2018language,dai2019transformer}.

\section{The Clinical Notes Generation Task}

To establish the merit of synthetic clinical notes generated by statistical models, we propose a task setup that consists of: (1)~real de-identified clinical notes datasets used to train models, which in turn generate synthetic notes; (2)~privacy measures used to estimate the privacy preservation properties of the synthetic notes; and (3)~utility benchmarks used to estimate the usefulness of the notes. To be considered successful, a model needs to score well both on privacy and utility measures.

\subsection{Original Clinical Notes Data}

As our source for composing the real clinical notes datasets, we used MIMIC-III (v1.4) \cite{johnson2016mimic}, a large de-identified database that comprises nearly 60,000 hospital admissions for 38,645 adult patients. Despite having been stripped of patient identifiers, 
MIMIC's records are available to researchers only under strict terms of use that include careful access restrictions and completion of sensitive data training\footnote{\url{https://mimic.physionet.org/gettingstarted/access/}} due to privacy concerns. 

Training language models is expensive in terms of time and compute power. It is a common practice \cite{merity2016pointer} to evaluate language models that were trained on both a small dataset that is relatively quick to train on and a medium-sized dataset which can demonstrate some benefits of scale while still being manageable. Therefore, within MIMIC-III, following \scite{dernoncourt2017identification}, we focused on the discharge summary notes due to their content diversity and richness in natural language text. Further, we followed the recently introduced WikiText-2 and WikiText-103 datasets \cite{merity2016pointer} to determine plausible size, splits and most of the preprocessing of our datasets. 
These datasets include text from Wikipedia articles and are commonly used to benchmark general-domain language models. We name our respective benchmarks, MedText-2 and MedText-103.

\begin{table*}[t]
\centering
\begin{tabular}{ll}
\begin{tabular}{| l | c | c | c |}
\hline
 & Train & Valid & Test \\
\hline
 &  \multicolumn{3}{ c |}{\textbf{\emph{MedText-2}}}  \\
\hline
Notes & 1280  & 128 & 128 \\
Words & 2,259,966 & 228,795 & 219,650 \\
\hline
Vocab &  \multicolumn{3}{ c |}{24,052}  \\
OOV &  \multicolumn{3}{ c |}{1.5\%}  \\
\hline
 &  \multicolumn{3}{ c |}{\textbf{\emph{MedText-103}}}  \\
\hline
Notes &  59,396 & 128 & 128 \\
Words & 103,590,422 & 228,795 & 219,650 \\
\hline
Vocab &  \multicolumn{3}{ c |}{135,220}  \\
OOV &  \multicolumn{3}{ c |}{0.3\%}  \\
\hline
\end{tabular}
&
\begin{tabular}{| l | c | c | c |}
\hline
 & Train & Valid & Test \\
\hline
 &  \multicolumn{3}{ c |}{\textbf{\emph{WikiText-2}}}  \\
\hline
Articles &  600 & 60 & 60 \\
Words & 2,088,628 &  217,646  & 245,569 \\
\hline
Vocab &  \multicolumn{3}{ c |}{33,278}  \\
OOV &  \multicolumn{3}{ c |}{2.6\%}  \\
\hline
 &  \multicolumn{3}{ c |}{\textbf{\emph{WikiText-103}}}  \\
\hline
Articles & 28,475  & 60 & 60 \\
Words & 103,227,021 & 217,646 & 245,569 \\
\hline
Vocab &  \multicolumn{3}{ c |}{267,735}  \\
OOV &  \multicolumn{3}{ c |}{0.4\%}  \\
\hline
\end{tabular}
\end{tabular}
\caption{MedText vs. WikiText dataset statistics
}
\label{tab:original_notes}
\end{table*}

To create the MedText datasets, we first extracted the full text of the discharge summary notes from the NOTEEVENTS table available from MIMIC-III. Since the text includes arbitrary line splits, presumably for formatting reasons, we merged lines and then performed sentence splitting and word tokenization using the NLP toolkit spaCy.\footnote{\url{https://spacy.io/}}
We then randomly sampled notes to create the MedText-2 and MedText-103 datasets. Each of these datasets was split into train/validation/test subsets, with MedText-2 and MedText-103 comprising approximately 2 and 103 million word train sets, respectively, and sharing the same $\sim$200K-word validation and test sets. Finally, we replaced all words with an occurrence count below~3 with an \emph{unk} token.\footnote{This was done separately for MedText-2 and MedText-103 resulting in a discrepancy between their validation/test sets in terms of the \emph{unk} tokens.}

Table \ref{tab:original_notes} describes more precise statistics of the resulting MedText datasets, compared to the respective WikiText datasets. As seen, compared to the WikiText datasets, which are nearly identical in terms of word counts, we note that MedText exhibits notably smaller vocabulary sizes (24K vs. 33K and 135K vs. 267K) and Out-Of-Vocabulary (OOV) rates (1.5\% vs. 2.6\% and 0.3\% vs. 0.4\%).
We hypothesize that this is one of the artifacts of MedText being more domain-specific than WikiText, as it is restricted only to discharge summary notes. To this end, we note that to the best of our knowledge, unlike the general domain where popular language modeling benchmarks, such as WikiText, PTB and WMT \cite{Chelba2013}, are commonly used, there are no equivalent benchmarks specific to the medical domain. Therefore, as an independent contribution, we propose MedText as such a benchmark.



\subsection{The Privacy Measure}
\label{subsec:privacy_measures}

As mentioned in the Background section, while traditional de-identification methods, such as deleting patient identifiers, are an essential pre-requisite to protecting the privacy of patient data, it is well understood that they are not sufficient to provide  strong privacy guarantees. To address this, we propose to share the output of statistical models that were trained to generate synthetic data based on real de-identified data. 
While this intuitively seems to increase privacy preservation compared to sharing the real data, it is still not necessarily sufficient, due to potential private information leakage from such models.

To quantify the risk involved in sharing synthetic clinical notes, we propose to use an empirical measure of private information leakage. This measure is meant to serve two purposes: (1) help drive the development of synthetic clinical notes generation methods that preserve privacy; and (2) inform decision makers regarding the concrete risk in releasing any given synthetic notes dataset.

Our proposed measure is adopted from the field of Differential Privacy (DP).
Recently, \scite{long2017towards} proposed an empirical differential privacy measure, called Differential Training Privacy~(DTP).
Unlike DP guarantees, which are analyzed theoretically and apply only to specific models designed for DP, DTP is a local property of any model and a concrete training set. It can be derived by means of empirical computation to any trained model regardless of whether it has theoretical DP guarantees, and provides an estimate of the privacy risks associated with sharing the outputs of that concrete trained model. 
In this work, we base our privacy measures on the \emph{Pointwise Differential Training Privacy} (PDTP) metric \cite{long2017towards}, a more computationally efficient variant of DTP:

\begin{dmath}
PDTP_{M,T}(t) = 
\end{dmath}
\vspace{-5 pt}
$$\max_{y \in Y} (|\log p_{M(T)}(y|t) - \log p_{M(T \setminus \{t\})}(y|t)|)$$

\noindent for a classification model $M$, a set of possible class predictions $Y$, a training set $T$, and a specific target record $t \in T$ for which the risk is measured. The rationale for this measure is that to protect the privacy of $t$, the difference in the predictions of a model trained with $t$ versus those of a model trained without it, should be as small as possible, and in particular when it comes to predictions made when the model is applied to $t$ itself. 

For the purpose of measuring privacy, we make the assumption that the model $M$ that was trained to generate the synthetic notes can be queried for the conditional probability $\log p_{M(T)}(w_i^c|w_{1..i-1}^c)$, where $w_i^c$ is the $i$-th word in clinical note $c$, which is our equivalent of a record. \footnote{If $M$ does not disclose this information, then the synthetic notes it generates could be used to train a language model $M'$ that does, as an approximation for $M$.} We note that unlike in the setting of \scite{long2017towards}, where a single class $y$ is predicted for each record, for synthetic notes, we can view every generated word $w_i^c$ in $c$ as a separate class prediction. Accordingly, we propose Sequential-PDTP:

\begin{dmath}
S-PDTP_{M,T}(c) = \max_{i \in 1..|c|} \Big(|\log p_{M(T)}(w_i^c|w_{1..i-1}^c) \\ - \log p_{M(T \setminus \{c\})}(w_i^c|w_{1..i-1}^c)|\Big)
\label{eq:spdtp_c}
\end{dmath}

\noindent S-PDTP estimates the privacy risk for clinical note $c$ as the largest absolute difference between the conditional probability predictions made by $M(T)$ and $M(T \setminus \{c\})$ for any of the words in $c$ given their preceding context.
Finally, our proposed privacy score for notes generated by a model $M$ trained on a benchmark dataset $T$, is the expected privacy risk, where a higher score indicates a higher expected risk:

\begin{dmath}
S-PDTP_{M,T} = \mathbb{E}_{c \in T}[S-PDTP_{M,T}(c)]
\end{dmath}

Intuitively, a high S-PDTP score means that the output of the trained model is sensitive to the presence of at least some individual records in its training set and therefore revealing that output may compromise the private information in those records.
In practice, since it is challenging computationally to train and test $|T|$ different models, we use an estimated measure based on a sample of 30 notes from~T.


\subsection{Utility Benchmarks}

We compare the utility of synthetic vs. real clinical notes by using them as training data in the following clinical NLP tasks. 

\subsubsection{Estimating lexical-semantic association}
As a measure of the quality of the lexical semantic information contained in clinical notes, we use them to train word2vec embeddings \cite{Mikolov_nips} with 300 dimensions and a 5-word window~\footnote{We used default word2vec hyperparameters, except for 10 negative samples and 10 iterations.}. 
Then, we evaluate these embeddings on the medical word similarity and relatedness benchmarks, UMNSRS-Sim and
UMNSRS-Rel \cite{pakhomov2010semantic,chiu2016train}.
These benchmarks comprise 566 and 587 word pairs, which were manually rated with a similarity and relatedness score, respectively.

To evaluate each set of embeddings, we compute its estimated similarity scores, as the cosine similarity between the embeddings of the words in each pair. Since our MedText datasets are domain-specific and not huge in size, our learned embeddings do not include a representation for many of the words in the UMNSRS benchmarks. Therefore, to ensure that we do have an embedding for every word included in the evaluation, we limit our datasets only to pairs, whose words occur at least 20 times and 30 times in MedText-2 and MedText-103, respectively. Accordingly, the number of pairs we use from UMNSRS-Sim/UMNSRS-Rel is 110/105 in the case of MedText-2 and 317/305 in the case of MedText-103. Finally, each set of embeddings is evaluated according to the Spearman's correlation between the pair rankings induced by the embeddings' scores and the one induced by the manual scores.

\subsubsection{Natural language inference (NLI)} 
We also probe the utility of clinical notes for performing natural language inference (NLI) -- a sentence level task. The task is to determine whether a given hypothesis sentence can be inferred from a given premise sentence. NLI, also known as recognizing textual entailment (RTE) \cite{dagan2013recognizing}, is a fundamental popular task in natural language understanding.

For our NLI task, we use MedNLI, the first clinical domain NLI dataset, recently released by \scite{romanov2018lessons}. The dataset includes sentence pairs with annotated relations that are used to train evaluated models. \scite{romanov2018lessons} report the performance of various neural network based models that typically benefit from the use of unsupervised pre-trained word embeddings. In our benchmark,    
%
%
we report the accuracy of their simple \emph{BOW model} (also called \emph{sum of words}) with input embeddings that are pre-trained on MedText clinical notes and kept fixed during the training with the MedNLI sentence pairs. The pre-trained embeddings used were the same as the ones used for the lexical-semantic association task. 
In all of our experiments, we used the implementation of \scite{romanov2018lessons} with its default hyperparameters .\footnote{\url{https://github.com/jgc128/mednli}}

\subsubsection{Recovering letter case information} 
Our third task goes beyond word embeddings, using clinical notes to train a recurrent neural network model end-to-end. 
More specifically, we use MedText to train letter casing (capitalization) models.
These models are trained based on parallel data comprising the original text and an all-lowered-case version of the same. Then, they are evaluated on their ability to recover casing for a test lower-cased text. The appealing aspect of this task is that the parallel data can be easily obtained in various languages and domains.

We note that sequential information is important in predicting the correct casing of words. The simplest example in English is that the first word of every sentence usually begins with a capital letter, but title casing, and ambiguous words in context (such as the word `bid' that may need to be mapped to `BID', i.e. 'twice-a-day', in the clinical prescription context), are other examples.
Arguably, for this reason, the state-of-the-art for this task is achieved by sequential character-RNN models \cite{susanto2016learning}. We use their implementation\footnote{\url{https://github.com/raymondhs/char-rnn-truecase}} with default hyperparameters for our evaluation.\footnote{We use their `small' model configuration for MedText-2 and `large' model configuration for MedText-103.} 
We use the dev and test splits of MedText 
to perform the letter case recovery task and report F1.

\section{Experiments}

In this section, we describe results obtained when using various models to perform the clinical notes generation task. 
We first generate synthetic clinical notes and evaluate their privacy properties. Then, assuming these notes were shared with another party we evaluate their utility to that party in training various clinical NLP models compared to that of the real notes.

\subsection{Compared Methods}

To generate the synthetic notes, we used primarily a standard LSTM language model implementation by PyTorch.
\footnote{\href{https://github.com/pytorch/examples/tree/master/word_language_model}{https://github.com/pytorch/examples/}}

We trained 2-layer LSTM models with  650 hidden-units on the train sets of MedText-2 and MedText-103, and tuned their hyperparameters based on validation perplexity.\footnote{For MedText-2, we trained for 20 epochs, beginning with a learning rate of 20 and reducing it by a factor of 4 after every epoch for which the validation loss did not go down compared to the previous epoch. For the much larger MedText-103, we trained for 2 epochs, beginning with a learning rate of 20 and reducing it by a factor of 1.2 every $\frac{1}{40}$ epoch if the validation loss did not go down by at least 0.1, but never going below a minimum learning rate of 0.1. In all runs, we used SGD with gradients clipped to 0.25, back-propagation-through-time 35 steps, a batch size of 20 and tied input and output embeddings.} 

To get more perspective on the efficacy of the LSTM models, we also trained a simple unigram baseline with Lidstone smoothing:

\begin{dmath}
p_{unigram}(w_i = u|w_{1..i-1}) = \frac{count(u)+1}{N + |V|} 
\end{dmath}

\noindent where $w_i$ is the word at position $i$, $N$ is the total number of words in the train set and $|V|$ is the size of the vocabulary. As can clearly be seen, this is a very naive model that generates words based on a smoothed unigram distribution, disregarding the context of the word in the note. Therefore, we expect that the utility of notes generated with this model would be low. However, on the other hand, since it captures much less information about the train data, we also expect it to have better privacy properties.

We then used the trained models to generate synthetic MedText-2-M and MedText-103-M datasets with identical word counts to the respective real note train datasets, and where $M$ denotes a generative model being used. To that end, we iteratively sampled a next token from the model's predicted conditional probability distribution and then fed that token as input back to the model. We used an empty line as an indication of an end-of-note, hence a collection of clinical notes is represented by the model as a seamless sequence of text.

We study the effect that using dropout regularization  \cite{srivastava2014dropout,zaremba2014recurrent} has on privacy and the tradeoffs between privacy and utility. Dropout, like other regularization methods, is a machine learning technique commonly applied to neural networks to minimize their prediction error on unseen data by reducing overfitting to the train data. It has also been shown that avoiding overfitting using regularization is helpful for protecting the privacy of the train data \cite{jain2015drop,shokri2017membership,yeom2017unintended}. Accordingly, we hypothesize that the higher dropout values used in our models are, the better the privacy scores would be. Utility, however, typically has a dropout optimum value over which it begins to degrade. 

\subsection{Qualitative Observations}

We sought feedback from a clinician on the quality of the generated synthetic discharge summary notes. A generated note comprises various relevant sections indicated by plain text headers. These sections are mostly in the right order with a typical order being: admission details, medical history, treatment, medications and finally, discharge details. The text of a section is mostly topically coherent with its header. For instance, the text generated for a medical history section often includes sentences mentioning medical problems. On the other hand, although local linguistic expressions and phrases typically make sense, continuity across consecutive sentences makes little clinical sense and many sentences are unclear due to incorrect grammar. 
A simple but obvious error is change of gender for the same patient (e.g. the pronoun `he' switches to `she').
A different example for short range language modeling problem is generation of incorrect terms like ``Hepatitis C deficiency".
The quality of a generated section is typically much better when it is backed by a structure as in a numbered list of medications. Yet, a notable problem here is that lists frequently have repeated entries (e.g. same symptom listed more than once). In conclusion, to a human eye, the synthetic notes are clearly distinct from real ones, yet from a topical and shallow linguistic perspective they do carry genuine properties of the original content. A sample snippet of a synthetic clinical note is shown in Figure \ref{fig:gen-sample}.

\begin{figure}[ht]
\begin{framed}
\small
Admission Date $ \colon $ \\
$ \langle $ deidentified $ \rangle $ \\ 
Discharge Date $ \colon $ \\
$ \langle $ deidentified $ \rangle $ \\
Date of Birth $ \colon $ \\
$ \langle $ deidentified $ \rangle $ Sex $ \colon $ \\
F \\
Service $ \colon $ \\
SURGERY \\
Allergies $ \colon $ \\ 
Patient recorded as having No Known Allergies to Drugs \\
Attending $ \colon $ \\
$ \langle $ deidentified $ \rangle $ \\ 
Chief Complaint $ \colon $ \\ 
Dyspnea \\
Major Surgical or Invasive Procedure $ \colon $ \\
Mitral Valve Repair \\
History of Present Illness $ \colon $ \\
Ms. $ \langle $ deidentified $ \rangle $ is a 53 year old female who presents after a large bleed rhythmically lag to 2 dose but the patient was brought to the Emergency Department where he underwent craniotomy with stenting of right foot under the LUL COPD and transferred to the OSH on $ \langle $ deidentified $ \rangle $ . \\
The patient will need a pigtail catheter to keep the sitter daily . 
\end{framed}
\caption{Sample snippet of a synthetic clinical note}
\label{fig:gen-sample}
\end{figure}

\subsection{Results}

\begin{table*}[t]
\centering
\begin{tabular}{ | c | c || c | c || c | c | c | c |}
\hline
note generation model & dropout & perplexity & privacy & similarity & relatedness & nli & case\\
\hline
   \rowcolor[gray]{.9}
 \multicolumn{8}{| c |}{\emph{MedText-2}}  \\
\hline
 \multicolumn{4}{| c ||}{Baseline: Real notes}  & .459 & .381 & .713 & .910 \\
   \rowcolor[gray]{.9}
 \multicolumn{8}{| c |}{\emph{MedText-2-M}}  \\
\hline

\multirow{5}{*}{LSTM} & 0.0  & 15.8 & 11.7 & .227 & .125 & .678 & .895 \\
 & 0.3  & 12.5 & 11.8 & & & & \\
 & 0.5  & 12.5 & 9.6 & .259 & .160 & .692 & .895\\
 & 0.7  & 15.4 & 7.5 & & & & \\
 & 0.8  & 20.3 & 6.6 & .146 & .016 & .699 & .883 \\
 \hline
 unigram & N/A & 702.4 & 0.9 & .027 & -.072 & .661 & .488 \\
 
\hline
\multicolumn{7}{ c }{}  \\
\end{tabular}

\begin{tabular}{ | c | c || c | c || c | c | c | c |}
\hline
note generation model & dropout & perplexity & privacy & similarity & relatedness & nli & case\\
\hline

   \rowcolor[gray]{.9}
 \multicolumn{8}{| c |}{\emph{MedText-103}}  \\
\hline
 \multicolumn{4}{| c ||}{Baseline: Real notes}  & .608 & .489 & .724 & .921 \\
 \hline
   \rowcolor[gray]{.9}
 \multicolumn{8}{| c |}{\emph{MedText-103-M}}  \\
\hline

 \multirow{3}{*}{LSTM} & 0.0  & 7.8 & 4.9 & .415 & .351 & .697 & .918\\
 & 0.2  &  8.4 & 4.0 & .401 & .337 & .702 & .915 \\
 & 0.5  & 10.2 & 3.7 & .315 & .271 & .713 & .910 \\
 \hline
unigram & N/A & 803.5 & 0.3 & .094 & .170 & .644 & .469 \\

 \hline
\end{tabular}
\caption{Experimental results with the real MedText and synthetic MedText-M. `dropout' is the dropout value used to train different LSTM models on MedText and then generate the respective synthetic MedText-M datasets (0.0 means no dropout applied); `perplexity' is the perplexity obtained on the real MedText validation set for each note generation model $M$; `privacy' is our privacy measure ($S-PDTP_{M,T}$ for every $M$, where $T$ is MedText); `similarity'/`relatedness' are UMNSRS word similarity/relatedness correlation results obtained using word embeddings trained on MedText and MedText-M; `nli' is the accuracy obtained on the MedNLI test set using different MedText pre-trained word embeddings; and `case' is the case restoration F1 measure. 
}
\label{tab:pp}

\end{table*}

Table \ref{tab:pp} shows the results we get when training the LSTM language models with varied dropout values. Starting with perplexity, we see that generally we achieve notably lower (better) perplexities on MedText, compared to results with LSTM on WikiText, which are around 100 for WikiText-2 and 50 for WikiText-103. \footnote{https://www.salesforce.com/products/einstein/ai-research/the-wikitext-dependency-language-modeling-dataset/} We hypothesize that this may be due to the highly domain-specific medical jargon and repeating note template characteristics that are presumably more predictable.
We also see that best perplexity results are achieved with dropout values around 0.3-0.5 for MedText-2, and 0.0 (i.e. no dropout) for MedText-103, compared to the 0.5 dropout rate commonly used in general-domain language modeling \cite{zaremba2014recurrent,merity2016pointer}. These differences reinforce our proposal of MedText as an interesting language modeling benchmark for medical texts. As a reference for future work, we report the perplexity results obtained on the test set data: 12.88 on MedText-2 (dropout = 0.5), and 8.15 on MedText-103 (dropout = 0.0).

Next, looking at privacy, we see that as predicted, more aggressive (higher) dropout values yield better (lower) privacy risk scores. We also see that privacy scores on the large MedText-103 are generally much better than the ones on the smaller MedText-2. This observation is intuitive in the sense that we would expect to generally get better privacy protection when any single personal clinical note is mixed with more, rather than fewer, notes in the train-set of a note-generating model.

For the utility evaluation, we chose three representative dropout values, for which we generated the MedText-M notes and compared them against the real MedText on the utility benchmarks. Looking at the results, we first see, as expected, that the performance with MedText-M is consistently lower than that with MedText, i.e. real notes are more useful than synthetic ones. However, the synthetic notes do seem to bear useful information. In particular, in the case of the letter case recovery task, they perform almost as well as the real ones.
We also see as suspected, that privacy usually comes at some expense of utility.

Finally, looking at the unigram baseline, we see as expected that perplexity and utility is by far worse than that achieved by the LSTM models, while privacy is much better. This is yet further evidence of the utility vs. privacy trade-off. We hope that future work could reveal better models that can get closer to the privacy protection values exhibited by the unigram model, while achieving utility, which is closer to that of the real notes.

\subsection{Analysis}

To better understand the factors determining our proposed privacy scores, we took a closer look at two note generating models, \emph{MedText-2-0} and \emph{MedText-103-0}, which are the models trained on MedText-2 and MedText-103, respectively, with dropout=0.0. First, we note that in 30 out of 30 and 25 out of 30 of the notes sampled to compute $S-PDTP_{M,T}(c)$ (Eq. \ref{eq:spdtp_c}) in MedText-2-0 and MedText-103-0, respectively, we observe that
$$
\log p_{M(T)}(w_j^c|w_{1..j-1}^c) > \log p_{M(T \setminus \{c\})}(w_j^c|w_{1..j-1}^c)
$$
where
\vspace{-5 pt}
$$
j = argmax_{i \in 1..|c|} \Big(|\log p_{M(T)}(w_i^c|w_{1..i-1}^c) $$ 
\vspace{-15pt}
$$- \log p_{M(T \setminus \{c\})}(w_i^c|w_{1..i-1}^c)|\Big)
$$
In other words, in the vast majority of the cases, the maximum differences in probability predictions are due to the model trained on train-set $T$, which includes note $c$, estimating a higher conditional probability to a word in $c$ than the one estimated by the model trained on $T \setminus \{c\}$. 
This can be expected, since $M(T)$ has seen all the text in $c$ during training, while $M(T \setminus \{c\})$ may or may not have seen similar texts.

Furthermore, when looking at the actual text positions $j$ that determine the privacy scores, we indeed see that the prediction differences that contribute to the privacy risk measure, are typically due to rare words and/or sequences of words in note $c$ that have no similar counterparts in $T \setminus \{c\}$. 
More specifically,
several of the cases where $\log p_{M(T \setminus \{c\})}(w_j^c|w_{1..j-1}^c) \ll \log p_{M(T)}(w_{j}^c|w_{1..j-1}^c)$ occur when:
(1)~A particular rare word $w_j^c$, such as \emph{cutdown}, appears only in a single clinical note $c$ and never in $T \setminus \{c\}$. This happens, for example, in $p($``cutdown'' $|$ ``Left popliteal''$)$;\footnote{POD stands for `postoperative day'} (2)~The rare word is at position $j-1$ as is \emph{Ketamine} in $p($``gtt'' $|$ ``On POD \# 2 Ketamine,''$)$; and (3)~The word $w_j^c$ is not rare, but usually does not appear right after the sequence $w_{1..j-1}$ as in $p($``mouth'' $|$ ``foaming at''$)$, where in $T \setminus \{c\}$ there is always a determiner or pronoun before the word \emph{mouth}, or $p($``pain'' $|$ ``mild left should''$)$, where \emph{should} is a typo of \emph{shoulder}.

These findings lead us to hypothesize that cases of PHI, such as full names of patients, inadvertently left in de-identified notes, might desirably increase the privacy risk measure output because of their rarity. This would be interesting to validate in future work.

For risk mitigation, we hypothesize that using pre-trained word embeddings including rare words and even more so, pre-training the language model on a larger public out-of-domain resource \cite{howard2018fine}, may help in reducing some of the above discrepancies between $p_{M(T \setminus \{c\})}$ and $p_{M(T)}$ and hence improve the overall privacy score of the models.

\section{Related Work}

Recently, \scite{choi2017generating} proposed \emph{medGAN}, a model for generating synthetic patient records that are safer to share than the real ones due to stronger privacy properties. However, unlike our work, their study is focused on discrete variable records and does not address the wealth of information embedded in natural language notes.

\scite{boag2018towards} created a corpus of synthetically-identified clinical notes with the purpose of using this resource to train de-identification models. Unlike our synthetic notes, their notes only populate the PHI instances with synthetic data (e.g. replacing  ``[**Patient Name**] visited [**Hospital**]'' with the randomly sampled names ``Mary Smith visited MGH.''

\section{Conclusions and Future Work}

We proposed synthetic clinical notes generation as means to promote open and collaborative medical NLP research. To have merit, the synthetic notes need to be useful and at the same time better preserve the privacy of patients. To track progress on this front, we suggested a privacy measure and a few utility benchmarks. Our experiments using neural language models demonstrate the potential and challenges of this approach, reveal the expected trade-offs between privacy and utility, and provide baselines for future work. 

Further work is required to extend the range of clinical NLP tasks that can benefit from the synthetic notes as well as increase the levels of privacy provided. \scite{mcmahan2017learning} introduced an LSTM neural language model with differential privacy guarantees that has just been publicly released.\footnote{\url{https://github.com/tensorflow/privacy/}} \scite{radford2018language} and \scite{dai2019transformer} recently showed impressive improvement in language modeling performance using the novel attention-based \emph{Transformer} architecture and larger model scales. These methods are example candidates for evaluation on our proposed clinical notes generation task. 
With sufficient progress, we hope that this line of research would lead to useful large synthetic clinical notes datasets that would be available more freely to a wider research community.

\section*{Acknowledgments}

We would like to thank Ken Barker and Vandana Mukherjee for supporting this project. We would also like to thank Thomas Steinke for helpful discussions.
\bibliography{main}

\begin{thebibliography}{47}
\expandafter\ifx\csname natexlab\endcsname\relax\def\natexlab#1{#1}\fi

\bibitem[{Abadi et~al.(2016)Abadi, Chu, Goodfellow, McMahan, Mironov, Talwar,
  and Zhang}]{abadi2016deep}
Martin Abadi, Andy Chu, Ian Goodfellow, H~Brendan McMahan, Ilya Mironov, Kunal
  Talwar, and Li~Zhang. 2016.
\newblock Deep learning with differential privacy.
\newblock In \emph{Proceedings of the 2016 ACM SIGSAC Conference on Computer
  and Communications Security}.

\bibitem[{Afzal et~al.(2018)Afzal, Mallipeddi, Sohn, Liu, Chaudhry, Scott,
  Kullo, and Arruda-Olson}]{afzal2018natural}
Naveed Afzal, Vishnu~Priya Mallipeddi, Sunghwan Sohn, Hongfang Liu, Rajeev
  Chaudhry, Christopher~G Scott, Iftikhar~J Kullo, and Adelaide~M Arruda-Olson.
  2018.
\newblock Natural language processing of clinical notes for identification of
  critical limb ischemia.
\newblock \emph{International Journal of Medical Informatics}, 111:83--89.

\bibitem[{Birman-Deych et~al.(2005)Birman-Deych, Waterman, Yan, Nilasena,
  Radford, and Gage}]{birman2005accuracy}
Elena Birman-Deych, Amy~D Waterman, Yan Yan, David~S Nilasena, Martha~J
  Radford, and Brian~F Gage. 2005.
\newblock Accuracy of icd-9-cm codes for identifying cardiovascular and stroke
  risk factors.
\newblock \emph{{Medical Care}}, pages 480--485.

\bibitem[{Boag et~al.(2016)Boag, Naumann, and Szolovits}]{boag2018towards}
Willie Boag, Tristan Naumann, and Peter Szolovits. 2016.
\newblock Towards the creation of a large corpus of synthetically-identified
  clinical notes.
\newblock In \emph{In Proceedings of Machine Learning for Health Workshop at
  NIPS}.

\bibitem[{Chelba et~al.(2014)Chelba, Mikolov, M.Schuster, Ge, Brants, Koehn,
  and Robinson}]{Chelba2013}
C.~Chelba, T.~Mikolov, M.Schuster, Q.~Ge, T.~Brants, P.~Koehn, and T.~Robinson.
  2014.
\newblock One billion word benchmark for measuring progress in statistical
  language modeling.
\newblock In \emph{Proceedings of INTERSPEECH}.

\bibitem[{Chen et~al.(2015)Chen, Tan, Liu, Lanchantin, Wan, Gales, and
  Woodland}]{chen2015recurrent}
Xie Chen, Tian Tan, Xunying Liu, Pierre Lanchantin, Moquan Wan, Mark~JF Gales,
  and Philip~C Woodland. 2015.
\newblock Recurrent neural network language model adaptation for multi-genre
  broadcast speech recognition.
\newblock In \emph{Sixteenth Annual Conference of the International Speech
  Communication Association}.

\bibitem[{Chiu et~al.(2016)Chiu, Crichton, Korhonen, and
  Pyysalo}]{chiu2016train}
Billy Chiu, Gamal Crichton, Anna Korhonen, and Sampo Pyysalo. 2016.
\newblock How to train good word embeddings for biomedical nlp.
\newblock In \emph{Proceedings of the 15th Workshop on Biomedical Natural
  Language Processing}.

\bibitem[{Choi et~al.(2016)Choi, Bahadori, Schuetz, Stewart, and
  Sun}]{choi2016doctor}
Edward Choi, Mohammad~Taha Bahadori, Andy Schuetz, Walter~F Stewart, and Jimeng
  Sun. 2016.
\newblock Doctor ai: Predicting clinical events via recurrent neural networks.
\newblock In \emph{Machine Learning for Healthcare Conference}, pages 301--318.

\bibitem[{Choi et~al.(2017)Choi, Biswal, Malin, Duke, Stewart, and
  Sun}]{choi2017generating}
Edward Choi, Siddharth Biswal, Bradley Malin, Jon Duke, Walter~F Stewart, and
  Jimeng Sun. 2017.
\newblock Generating multi-label discrete patient records using generative
  adversarial networks.
\newblock In \emph{Proceedings of Machine Learning for Healthcare Conference}.

\bibitem[{Dagan et~al.(2013)Dagan, Roth, Sammons, and
  Zanzotto}]{dagan2013recognizing}
Ido Dagan, Dan Roth, Mark Sammons, and Fabio~Massimo Zanzotto. 2013.
\newblock Recognizing textual entailment: Models and applications.
\newblock \emph{Synthesis Lectures on Human Language Technologies},
  6(4):1--220.

\bibitem[{Dai et~al.(2019)Dai, Yang, Yang, Cohen, Carbonell, Le, and
  Salakhutdinov}]{dai2019transformer}
Zihang Dai, Zhilin Yang, Yiming Yang, William~W Cohen, Jaime Carbonell, Quoc~V
  Le, and Ruslan Salakhutdinov. 2019.
\newblock Transformer-xl: Attentive language models beyond a fixed-length
  context.
\newblock \emph{arXiv preprint arXiv:1901.02860}.

\bibitem[{Dernoncourt et~al.(2017)Dernoncourt, Lee, Uzuner, and
  Szolovits}]{dernoncourt2017identification}
Franck Dernoncourt, Ji~Young Lee, Ozlem Uzuner, and Peter Szolovits. 2017.
\newblock De-identification of patient notes with recurrent neural networks.
\newblock \emph{Journal of the American Medical Informatics Association},
  24(3):596--606.

\bibitem[{Douglass et~al.(2004)Douglass, Clifford, Reisner, Moody, and
  Mark}]{douglass2004computer}
Margaret Douglass, Gari~D Clifford, Andrew Reisner, George~B Moody, and Roger~G
  Mark. 2004.
\newblock Computer-assisted de-identification of free text in the mimic ii
  database.
\newblock In \emph{{Computers in Cardiology, 2004}}, pages 341--344. IEEE.

\bibitem[{Dwork(2008)}]{dwork2008differential}
Cynthia Dwork. 2008.
\newblock Differential privacy: A survey of results.
\newblock In \emph{Proceedings of International Conference on Theory and
  Applications of Models of Computation}.

\bibitem[{El~Emam et~al.(2011)El~Emam, Jonker, Arbuckle, and
  Malin}]{el2011systematic}
Khaled El~Emam, Elizabeth Jonker, Luk Arbuckle, and Bradley Malin. 2011.
\newblock A systematic review of re-identification attacks on health data.
\newblock \emph{PloS One}, 6(12):e28071.

\bibitem[{Gkoulalas-Divanis et~al.(2014)Gkoulalas-Divanis, Loukides, and
  Sun}]{gkoulalas2014publishing}
Aris Gkoulalas-Divanis, Grigorios Loukides, and Jimeng Sun. 2014.
\newblock Publishing data from electronic health records while preserving
  privacy: A survey of algorithms.
\newblock \emph{Journal of biomedical informatics}, 50:4--19.

\bibitem[{Gobbel et~al.(2014)Gobbel, Garvin, Reeves, Cronin, Heavirland,
  Williams, Weaver, Jayaramaraja, Giuse, Speroff et~al.}]{gobbel2014assisted}
Glenn~T Gobbel, Jennifer Garvin, Ruth Reeves, Robert~M Cronin, Julia
  Heavirland, Jenifer Williams, Allison Weaver, Shrimalini Jayaramaraja, Dario
  Giuse, Theodore Speroff, et~al. 2014.
\newblock Assisted annotation of medical free text using raptat.
\newblock \emph{{Journal of the American Medical Informatics Association}},
  21(5):833--841.

\bibitem[{Hanauer et~al.(2013)Hanauer, Aberdeen, Bayer, Wellner, Clark, Zheng,
  and Hirschman}]{hanauer2013bootstrapping}
David Hanauer, John Aberdeen, Samuel Bayer, Benjamin Wellner, Cheryl Clark, Kai
  Zheng, and Lynette Hirschman. 2013.
\newblock Bootstrapping a de-identification system for narrative patient
  records: cost-performance tradeoffs.
\newblock \emph{{International Journal of Medical Informatics}},
  82(9):821--831.

\bibitem[{Howard and Ruder(2018)}]{howard2018fine}
Jeremy Howard and Sebastian Ruder. 2018.
\newblock Fine-tuned language models for text classification.
\newblock In \emph{Proceedings of ACL}.

\bibitem[{Jain et~al.(2015)Jain, Kulkarni, Thakurta, and
  Williams}]{jain2015drop}
Prateek Jain, Vivek Kulkarni, Abhradeep Thakurta, and Oliver Williams. 2015.
\newblock To drop or not to drop: Robustness, consistency and differential
  privacy properties of dropout.
\newblock \emph{arXiv preprint arXiv:1503.02031}.

\bibitem[{Johnson et~al.(2016)Johnson, Pollard, Shen, Li-wei, Feng, Ghassemi,
  Moody, Szolovits, Celi, and Mark}]{johnson2016mimic}
Alistair~EW Johnson, Tom~J Pollard, Lu~Shen, H~Lehman Li-wei, Mengling Feng,
  Mohammad Ghassemi, Benjamin Moody, Peter Szolovits, Leo~Anthony Celi, and
  Roger~G Mark. 2016.
\newblock {MIMIC-III, a freely accessible critical care database}.
\newblock \emph{Scientific data}, 3:160035.

\bibitem[{Jozefowicz et~al.(2016)Jozefowicz, Vinyals, Schuster, Shazeer, and
  Wu}]{jozefowicz2016exploring}
R.~Jozefowicz, O.~Vinyals, M.~Schuster, N.~Shazeer, and Y.~Wu. 2016.
\newblock Exploring the limits of language modeling.
\newblock \emph{arXiv preprint arXiv:1602.02410}.

\bibitem[{K{\"o}pcke et~al.(2013)K{\"o}pcke, Trinczek, Majeed, Schreiweis,
  Wenk, Leusch, Ganslandt, Ohmann, Bergh, R{\"o}hrig
  et~al.}]{kopcke2013evaluation}
Felix K{\"o}pcke, Benjamin Trinczek, Raphael~W Majeed, Bj{\"o}rn Schreiweis,
  Joachim Wenk, Thomas Leusch, Thomas Ganslandt, Christian Ohmann, Bj{\"o}rn
  Bergh, Rainer R{\"o}hrig, et~al. 2013.
\newblock Evaluation of data completeness in the electronic health record for
  the purpose of patient recruitment into clinical trials: a retrospective
  analysis of element presence.
\newblock \emph{{BMC medical informatics and decision making}}, 13(1):37.

\bibitem[{Liu et~al.(2018)Liu, Zhang, and Razavian}]{liu2018deep}
Jingshu Liu, Zachariah Zhang, and Narges Razavian. 2018.
\newblock Deep ehr: Chronic disease prediction using medical notes.
\newblock \emph{arXiv preprint arXiv:1808.04928}.

\bibitem[{Liu et~al.(2017)Liu, Tang, Wang, and Chen}]{liu2017identification}
Zengjian Liu, Buzhou Tang, Xiaolong Wang, and Qingcai Chen. 2017.
\newblock De-identification of clinical notes via recurrent neural network and
  conditional random field.
\newblock \emph{{Journal of Biomedical Informatics}}, 75:S34--S42.

\bibitem[{Long et~al.(2017)Long, Bindschaedler, and Gunter}]{long2017towards}
Yunhui Long, Vincent Bindschaedler, and Carl~A Gunter. 2017.
\newblock Towards measuring membership privacy.
\newblock \emph{arXiv preprint arXiv:1712.09136}.

\bibitem[{Luong et~al.(2015)Luong, Kayser, and Manning}]{luong2015deep}
Thang Luong, Michael Kayser, and Christopher~D Manning. 2015.
\newblock Deep neural language models for machine translation.
\newblock In \emph{Proceedings of the Nineteenth Conference on Computational
  Natural Language Learning}.

\bibitem[{McMahan et~al.(2018)McMahan, Ramage, Talwar, and
  Zhang}]{mcmahan2017learning}
H~Brendan McMahan, Daniel Ramage, Kunal Talwar, and Li~Zhang. 2018.
\newblock Learning differentially private language models without losing
  accuracy.
\newblock In \emph{Proceedings of ICLR}.

\bibitem[{Merity et~al.(2017)Merity, Xiong, Bradbury, and
  Socher}]{merity2016pointer}
S.~Merity, C.~Xiong, J.~Bradbury, and R.~Socher. 2017.
\newblock Pointer sentinel mixture models.
\newblock In \emph{Proceedings of ICLR}.

\bibitem[{Mikolov et~al.(2013)Mikolov, Sutskever, Chen, Corrado, and
  Dean}]{Mikolov_nips}
T.~Mikolov, I.~Sutskever, K.~Chen, G.~Corrado, and J.~Dean. 2013.
\newblock Distributed representations of words and phrases and their
  compositionality.
\newblock In \emph{Advances in Neural Information Processing Systems}.

\bibitem[{Morrison et~al.(2009)Morrison, Li, Lai, and
  Hripcsak}]{morrison2009repurposing}
Frances~P Morrison, Li~Li, Albert~M Lai, and George Hripcsak. 2009.
\newblock Repurposing the clinical record: can an existing natural language
  processing system de-identify clinical notes?
\newblock \emph{{Journal of the American Medical Informatics Association}},
  16(1):37--39.

\bibitem[{Neamatullah et~al.(2008)Neamatullah, Douglass, Li-wei, Reisner,
  Villarroel, Long, Szolovits, Moody, Mark, and
  Clifford}]{neamatullah2008automated}
Ishna Neamatullah, Margaret~M Douglass, H~Lehman Li-wei, Andrew Reisner,
  Mauricio Villarroel, William~J Long, Peter Szolovits, George~B Moody, Roger~G
  Mark, and Gari~D Clifford. 2008.
\newblock Automated de-identification of free-text medical records.
\newblock \emph{{BMC Medical Informatics and Decision Making}}, 8(1):32.

\bibitem[{Ohm(2009)}]{ohm2009broken}
Paul Ohm. 2009.
\newblock Broken promises of privacy: Responding to the surprising failure of
  anonymization.
\newblock \emph{Ucla L. Rev.}, 57:1701.

\bibitem[{Pakhomov et~al.(2010)Pakhomov, McInnes, Adam, Liu, Pedersen, and
  Melton}]{pakhomov2010semantic}
Serguei Pakhomov, Bridget McInnes, Terrence Adam, Ying Liu, Ted Pedersen, and
  Genevieve~B Melton. 2010.
\newblock Semantic similarity and relatedness between clinical terms: an
  experimental study.
\newblock In \emph{Proceedings of AMIA}.

\bibitem[{Papernot et~al.(2018)Papernot, Song, Mironov, Raghunathan, Talwar,
  and Erlingsson}]{papernot2018scalable}
Nicolas Papernot, Shuang Song, Ilya Mironov, Ananth Raghunathan, Kunal Talwar,
  and {\'U}lfar Erlingsson. 2018.
\newblock Scalable private learning with pate.
\newblock In \emph{Proceedings of ICLR}.

\bibitem[{Radford et~al.(2018)Radford, Wu, Child, Luan, Amodei, and
  Sutskever}]{radford2018language}
Alec Radford, Jeffrey Wu, Rewon Child, David Luan, Dario Amodei, and Ilya
  Sutskever. 2018.
\newblock Language models are unsupervised multitask learners.
\newblock Technical report, Technical report, OpenAi.

\bibitem[{Raghavan et~al.(2014)Raghavan, Chen, Fosler-Lussier, and
  Lai}]{raghavan2014essential}
Preethi Raghavan, James~L Chen, Eric Fosler-Lussier, and Albert~M Lai. 2014.
\newblock How essential are unstructured clinical narratives and information
  fusion to clinical trial recruitment?
\newblock In \emph{{Proceedings of AMIA Summits on Translational Science}},
  volume 2014. American Medical Informatics Association.

\bibitem[{Romanov and Shivade(2018)}]{romanov2018lessons}
Alexey Romanov and Chaitanya Shivade. 2018.
\newblock Lessons from natural language inference in the clinical domain.
\newblock In \emph{Proceedings of EMNLP}.

\bibitem[{Rosenbloom et~al.(2011)Rosenbloom, Denny, Xu, Lorenzi, Stead, and
  Johnson}]{rosenbloom2011data}
S~Trent Rosenbloom, Joshua~C Denny, Hua Xu, Nancy Lorenzi, William~W Stead, and
  Kevin~B Johnson. 2011.
\newblock Data from clinical notes: a perspective on the tension between
  structure and flexible documentation.
\newblock \emph{{Journal of the American Medical Informatics Association}},
  18(2):181--186.

\bibitem[{Shokri et~al.(2017)Shokri, Stronati, Song, and
  Shmatikov}]{shokri2017membership}
Reza Shokri, Marco Stronati, Congzheng Song, and Vitaly Shmatikov. 2017.
\newblock Membership inference attacks against machine learning models.
\newblock In \emph{Proceedings of IEEE Symposium on Security and Privacy}.

\bibitem[{Singh et~al.(2004)Singh, Holmgren, and
  Noorbaloochi}]{singh2004accuracy}
Jasvinder~A Singh, Aaron~R Holmgren, and Siamak Noorbaloochi. 2004.
\newblock Accuracy of veterans administration databases for a diagnosis of
  rheumatoid arthritis.
\newblock \emph{{Arthritis Care \& Research}}, 51(6):952--957.

\bibitem[{Srivastava et~al.(2014)Srivastava, Hinton, Krizhevsky, Sutskever, and
  Salakhutdinov}]{srivastava2014dropout}
Nitish Srivastava, Geoffrey Hinton, Alex Krizhevsky, Ilya Sutskever, and Ruslan
  Salakhutdinov. 2014.
\newblock Dropout: A simple way to prevent neural networks from overfitting.
\newblock \emph{The Journal of Machine Learning Research}, 15(1):1929--1958.

\bibitem[{Stubbs et~al.(2015)Stubbs, Kotfila, and Uzuner}]{stubbs2015automated}
Amber Stubbs, Christopher Kotfila, and {\"O}zlem Uzuner. 2015.
\newblock Automated systems for the de-identification of longitudinal clinical
  narratives: Overview of 2014 i2b2/uthealth shared task track 1.
\newblock \emph{{Journal of Biomedical Informatics}}, 58:S11--S19.

\bibitem[{Susanto et~al.(2016)Susanto, Chieu, and Lu}]{susanto2016learning}
Raymond~Hendy Susanto, Hai~Leong Chieu, and Wei Lu. 2016.
\newblock Learning to capitalize with character-level recurrent neural
  networks: An empirical study.
\newblock In \emph{Proceedings of EMNLP}.

\bibitem[{Sutskever et~al.(2011)Sutskever, Martens, and
  Hinton}]{sutskever2011generating}
Ilya Sutskever, James Martens, and Geoffrey~E Hinton. 2011.
\newblock Generating text with recurrent neural networks.
\newblock In \emph{Proceedings of ICML}.

\bibitem[{Yeom et~al.(2017)Yeom, Fredrikson, and Jha}]{yeom2017unintended}
Samuel Yeom, Matt Fredrikson, and Somesh Jha. 2017.
\newblock Privacy risk in machine learning: Analyzing the connection to
  overfitting.
\newblock In \emph{Procedings of the IEEE Computer Security Foundations
  Symposium}.

\bibitem[{Zaremba et~al.(2014)Zaremba, Sutskever, and
  Vinyals}]{zaremba2014recurrent}
Wojciech Zaremba, Ilya Sutskever, and Oriol Vinyals. 2014.
\newblock Recurrent neural network regularization.
\newblock \emph{arXiv preprint arXiv:1409.2329}.

\end{thebibliography}
\bibliographystyle{acl_natbib}

\end{document}